\documentclass{article}

\usepackage[preprint]{corl_2022} % Use this for the initial submission.
\usepackage{graphicx}
\usepackage{amssymb}
\usepackage{amsmath}
\DeclareMathOperator*{\argmax}{arg\,max}
\usepackage{enumerate}
\usepackage{float}
\usepackage{subfig}
\floatstyle{plaintop}
\restylefloat{table}
\usepackage[shortlabels]{enumitem}
\usepackage{multirow}
\usepackage{xfrac}

\title{Safe Inverse Reinforcement Learning via Control Barrier Function}

\author{
  Yue Yang, Letian Chen, Matthew Gombolay\\
  School of Interactive Computing \\
  Georgia Institute of Technology \\
  Atlanta, GA, 30332\\
  \texttt{\{yyang941, letian.chen, matthew.gombolay\}@gatech.edu} \\
}

\begin{document}

\maketitle

%===============================================================================
\vspace{-7pt}
\begin{abstract}
    Learning from Demonstration (LfD) is a powerful method for enabling robots to perform novel tasks as it is often more tractable for a non-roboticist end-user to demonstrate the desired skill and for the robot to efficiently learn from the associated data than for a human to engineer a reward function for the robot to learn the skill via reinforcement learning (RL). Safety issues arise in modern LfD techniques, e.g., Inverse Reinforcement Learning (IRL), just as they do for RL; yet, safe learning in LfD has received little attention. In the context of agile robots, safety is especially vital due to the possibility of robot-environment collision, robot-human collision, and damage to the robot. In this paper, we propose a safe IRL framework, CBFIRL, that leverages the Control Barrier Function (CBF) to enhance the safety of the IRL policy. The core idea of CBFIRL is to combine a loss function inspired by CBF requirements with the objective in an IRL method, both of which are jointly optimized via gradient descent. In the experiments, we show our framework performs safer compared to IRL methods without CBF, that is $\sim15\%$ and $\sim20\%$ improvement for two levels of difficulty of a 2D racecar domain and $\sim 50\%$ improvement for a 3D drone domain. 
\end{abstract}

% Two or three meaningful keywords should be added here

%===============================================================================

\section{Introduction}
\vspace{-7pt}
% 1. why lfd important for agile robots (could borrow from pingpong, leg robot papers)
% 1.1 agile robots use rl methods 1.2 agile robot use lfd
Agility is an indispensable feature for robots applied in manufacturing or everyday life~\citep{kootbally2018enabling, yuan2021human} because the physical space can change very fast and the robots need to react quickly. Recent advances in robot learning have offered the potential to improve the agility of various robots, including high-speed cars~\citep{chae2017autonomous, isele2018navigating, yang2017feature, wang2018deep}, drones~\citep{pham2018cooperative, akhloufi2019drones}, legged robots~\citep{heess2017emergence, peng2018deepmimic}, and robots in sports~\citep{liu2018learning, chen2020learning}. Reinforcement learning (RL) is a ubiquitous approach to robot learning for developing high-performance controllers for robots. Although various RL-based methods have shown promising results in both simulation and real robots, the design of reward functions that elicit desired behaviors could still be laborious and time-consuming~\citep{peng2020learning}. Also, agents trained with RL can behave unnaturally~\citep{peng2020learning}. Although it's hard to design controllers, humans can demonstrate robots for agile control (e.g., racecar driving and drone flying). As such, Learning from Demonstration (LfD), a field empowering end-users to program robots by demonstrations instead of a computing language~\citep{peng2020learning, ravichandar2020recent, pan2017agile, chen2020joint}, can help address the issues by learning from experts and work in a more sample-efficient way. 

% These issues motivate researchers to adopt Learning from Demonstration (LfD) techniques, a field empowering end-users to program robots by demonstrations instead of a computing language~\citep{peng2020learning, ravichandar2020recent, pan2017agile, chen2020joint}. 

% 2. safety issue in lfd method. 
Inverse Reinforcement Learning (IRL)~\citep{abbeel2004apprenticeship} is a technique in LfD research that aims to infer a demonstrator's underlying objective function (i.e., reward) from demonstrations. However, the safety of IRL approaches is yet to be explored. Previous works in safe IRL~\cite{brown2020safe,brown2018efficient,brown2020better,brown2017toward, brown2019deep} mainly focus on adding high-confidence bounds on the learned policy's performance, which is an indirect approach to promote safety and dangerous cases can still happen as the underlying objective function can guide the agent into danger. Therefore, a direct approach to avoid dangerous configurations in IRL, instead of hoping for safety in a performance-focused way, is needed. 

% 3. cbf is good way to tackle safe issues, and cbf is popular in rl (some related works). 
To address safety in RL~\citep{marvi2021safe, choi2020reinforcement, ma2021model, qin2021learning}, researchers have leveraged the control barrier function (CBF)~\citep{ames2014control, ames2019control}, which is a method to synthesize a control policy that maintains the system within a safe set of states. Although CBF can help directly avoid dangerous cases in RL, to the best of our knowledge, there's no work incorporating CBFs with IRL algorithms to mitigate possible safety issues. The most related works focus on synthesizing CBF from data (e.g., expert demonstrations) and combine the CBF with hand-designed control methods~\citep{lindemann2020learning, robey2020learning, srinivasan2020synthesis}. However, these methods fail to show how to make use of the synthesized CBF to enforce the learned policy in IRL safer.

In this paper, we propose a framework named CBFIRL where the CBF, approximated by a neural network, is learned and utilized to enforce the learned policy in IRL to take safe actions by optimizing a joint loss. We present empirical results over two simulated agile robot control tasks and find the proposed CBFIRL has $\sim15\%$ and $\sim20\%$ less collision for two levels of difficulty of one 2D racecar domain and $\sim 50\%$ less collision for one 3D drone domain than just IRL.

%   Quadratic Program (QP) is a typical method for solving the optimization problem~\citep{ames2019control}. However, the problem could become intractable when the system becomes complicated and function classes are not rich, which could oftentimes happen. 

%===============================================================================
\section{Preliminaries}
\label{sec:preliminaries}
\vspace{-7pt}
% In this section, we will introduce three fundamental building blocks of CBFIRL: Markov Decision Process, Inverse Reinforcement Learning, and Control Barrier Function.

\paragraph{Markov Decision Process --}
% where state $s\in \mathcal{S}$ and action $a\in \mathcal{A}$. 
We model the environment as a Markov Decision Process (MDP) $\mathcal{M}$~\citep{white1993survey}, which is defined by $\langle\mathcal{S}, \mathcal{A}, R, T, \gamma, \rho_0\rangle$. $\mathcal{S}$ and $\mathcal{A}$ denote the state space and action space, respectively. $R: \mathcal{S}\rightarrow \mathbb{R}$ is the reward function that tells the reward for a given state. $T : \mathcal{S} \times \mathcal{A} \rightarrow \mathcal{S}$ represents a deterministic transition function that gives the next state $s'$ after applying the action $a$ to current state $s$. $\gamma \in (0, 1)$ is the temporal discount factor. $\rho_0 : \mathcal{S} \rightarrow \mathbb{R}$ denotes the initial state distribution. A policy $\pi : \mathcal{S} \times \mathcal{A} \rightarrow \mathbb{R}$ is a mapping from states to probabilities over actions. We could generate a trajectory $\tau = <s_0, a_0, r_0, \cdots, s_t, a_t, r_t, \cdots>$ by executing the policy within the environment. The expected discounted return of one policy could be calculated by $J(\pi) = \mathbb{E}_{\tau \sim \pi}\left[\sum^\infty_{t=0}\gamma^t R(s_t)\right]$. The objective for RL is to find the optimal policy, $\pi^* = \argmax_{\pi} J(\pi)$.

\paragraph{Inverse Reinforcement Learning --}	
Inverse reinforcement learning (IRL) considers an MDP sans reward function and infers a reward function $\hat{R}$ from a set of demonstration trajectories, $\mathcal{D} = \{\tau_1, \tau_2, \cdots, \tau_N\}$. Our method is based on adversarial inverse reinforcement learning (AIRL)~\citep{fu2017learning}. AIRL consists of a generator (i.e., a policy) to imitate the demonstrator and a discriminator to distinguish the generator's behavior from that of the demonstrator. The discriminator is defined as $D_\theta = \sfrac{e^{\{\hat{f}_\theta(s, a)\}}}{(e^{\{\hat{f}_\theta(s, a)\}} + \pi_\phi(a|s))}$, where $\hat{f}_\theta(s, a)$ is the inferred advantage function and $\pi_\phi(a|s)$ is the learned policy parameterized by $\phi$. The discriminator is trained to minimize a binary cross entropy loss, $\mathcal{L}_D$. The generator policy $\pi_{\phi}(a|s)$ is trained by optimizing the policy loss, $\mathcal{L}_{policy} = \max J(\pi)$, and to maximize the recovered reward function.

% with policy gradient to maximize the recovered reward function $\^{R}$ and the demonstration policy will be recovered when reaching the global optimal point. 
% \begin{equation}
%     \label{eq:ld}
%     \mathcal{L}_D = -\mathbb{E}_{\tau \sim \mathcal{D}, (s, a)\sim\tau}[log D(s, a)] - \mathbb{E}_{\tau \sim \pi_\phi, (s, a)\sim\tau}[log(1 - D(s, a))]
% \end{equation}

\paragraph{Control Barrier Function -- } 
% The CBF~\citep{ames2014control} is an approach to achieve the property of invariance and therefore ensure safety. 
Let $\mathcal{S}_s \subset \mathcal{S}$ be the safe states set, $\mathcal{S}_d = \mathcal{S}\backslash\mathcal{S}_s$ be the dangerous states set, and $\mathcal{S}_0$ be the set of initial states. A control barrier function, $h$, needs to satisfy the three requirements~\citep{ames2019control, luo2021learning}:
\textbf{R1}: $\forall s \in \mathcal{S}_0$, $h(s) \geq 0$; 
\textbf{R2}: $\forall s \in \mathcal{S}_d$, $h(s) < 0$; and
\textbf{R3}: $\forall s \in \{s | h(s) \geq 0\}$, $ (h(T(s, \pi_\phi(s))) - h(s)) / \varDelta t + \alpha(h(s)) \geq 0$. 
Here, $\alpha(\cdot)$ is a class-$\mathcal{K}$ function (i.e., $\alpha(\cdot)$ is strictly increasing and $\alpha(0)=0$). \textbf{R1} and \textbf{R3} ensure trajectories to stay inside the superlevel set $\mathcal{C}_h = \{s \in \mathcal{S}: h(s) \geq 0\}$. \textbf{R2} guarantees that unsafe states will never be visited under the policy $\pi_\phi$. In order to obtain a safe policy $\pi_\phi(\cdot)$ and an $h(\cdot)$ to meet the three requirements, we formulate a similar optimization objective as~\citet{qin2021learning}. We denote $\mathcal{P}$ and $\mathcal{H}$ as the function classes for $\pi_\phi(\cdot)$ and $h(\cdot)$, and $\mathcal{T}$ as the set of all trajectories. We assume initial states are safe, i.e. $\forall s \in \mathcal{S}_s$, $h(s) \geq 0$. We then define the function $y: \mathcal{H} \times \mathcal{P} \times \mathcal{T} \rightarrow \mathbb{R}$ as given by Equation~\ref{eq:y_func}.
\begin{equation}
    \label{eq:y_func}
    y(h, \pi_\phi, \tau) := min\{\inf_{s \in \mathcal{S}_s} h(s), \inf_{s \in \mathcal{S}_d} -h(s), \inf_{\{s|h(s)\geq0\}\cap \tau}(h(T(s, \pi_\phi(s))) - h(s)) / \varDelta t + \alpha(h) \geq 0\}
\end{equation}
\textbf{R1}-\textbf{R3} are satisfied when we find $h(\cdot)$ and $\pi_\phi(\cdot)$ such that $y(h, \pi_\phi, \tau) > 0$ for $\forall \tau \in \mathcal{T}$. Thus, The optimization objective is given by Equation~\ref{eq:optimization}.
\begin{equation}
    \label{eq:optimization}
    \text{Find } h(\cdot) \in \mathcal{H} \text{ and } \pi_\phi(\cdot) \in \mathcal{P}, \qquad s.t. \qquad\forall \tau \in \mathcal{T},  y(h, \pi_\phi, \tau) > 0
\end{equation}

% where $\epsilon > 0$ is a margin, which could provide a higher confidence to satisfy the requirements.

% $w_{crr} = w_0^{1 + \frac{r_0 - r_{crr}}{\epsilon}}$
% \begin{equation}
%   w_{crr} =
%     \begin{cases}
%       w_0 & \text{if $r_0 - r_{crr} < \epsilon$}\\
%       0 & \text{otherwise}
%     \end{cases}       
% \end{equation}

%===============================================================================
\section{Method}
\label{sec:method}
\vspace{-7pt}

\label{sec:m1}
\begin{figure}[t]
    \centering
    \includegraphics[width=0.9\textwidth]{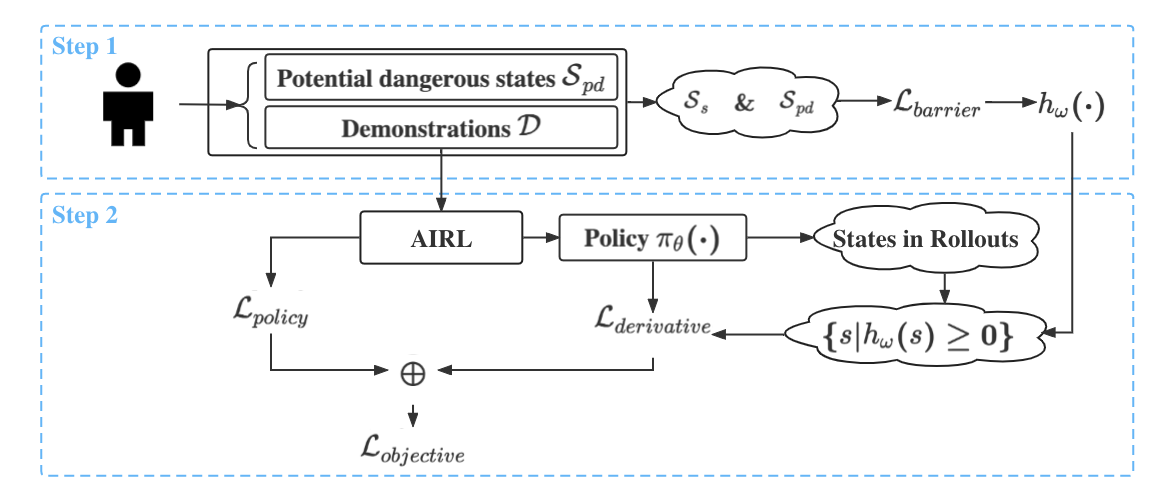}
 \caption{This figure depicts the architecture of CBFIRL.}
    \label{fig:arch}
    \vspace{-8pt}
\end{figure}

As shown in Figure~\ref{fig:arch}, we combine CBF with AIRL in two steps: \textbf{Step 1)} Request demonstrations from users together with potentially dangerous states (Section~\ref{sec:data}), which are then used to formulate a barrier loss function $\mathcal{L}_{barrier}$ to learn $h(\cdot)$ satisfying \textbf{R1} and \textbf{R2}; \textbf{Step 2)} leverage $h(\cdot)$ to formulate a derivative loss function $\mathcal{L}_{derivative}$ to synthesize a safer policy $\pi_\phi(\cdot)$ meeting \textbf{R3}.
For Step 1, we formulate the $\mathcal{L}_{barrier}$ as shown in Equation~\ref{eq:loss_barrier} based on Equation~\ref{eq:y_func}, where $h_\omega(\cdot)$ is a neural network parameterized by $\omega$. Terms in Equation~\ref{eq:loss_barrier} correspond to \textbf{R1} and \textbf{R2}. Intuitively, minimizing the $\mathcal{L}_{barrier}$ provides an $h_\omega(\cdot)$ that could discriminate safe states from dangerous ones. We collect the set $\mathcal{S}_s$ and $\hat{\mathcal{S}}_{pd}$ as described below in Section~\ref{sec:data}. 

\vspace{-10pt}
\begin{equation}
    \label{eq:loss_barrier}
    \mathcal{L}_{barrier} = \sum_{s\in\mathcal{S}_s}max(- h_\omega(s), 0) + \sum_{s\in\hat{\mathcal{S}}_{pd}}max(h_\omega(s), 0)
\end{equation}
\vspace{-10pt}

% frozen. The AIRL will be pre-trained to converge and provide a policy $\pi_\phi(\cdot)$. With the pre-optimized $h(\cdot)$, the set $\{s|h(s)\geq0\}$ is generated from the policy rollouts while keeping training AIRL. 

For Step 2, we formulate the $\mathcal{L}_{derivative}$ as shown in Equation~\ref{eq:loss_deriv}, where the policy $\pi_\phi(\cdot)$ is a neural network parameterized by $\phi$. 
\begin{equation}
    \label{eq:loss_deriv}
    \mathcal{L}_{derivative} = \sum_{s\in\{s|h_\omega(s)\geq0\}}max(- (h_\omega(T(s, \pi_\phi(s))) - h_\omega(s)) / \varDelta t - \alpha(h_\omega(s)), 0)
\end{equation}
For the class-$\mathcal{K}$ function, we use a linear function $\alpha(h(s)) = \lambda h(s)$. Minimizing the $\mathcal{L}_{derivative}$ enforces the $\pi_\phi(\cdot)$ to generate actions that satisfy requirement~\textbf{R3}, which provides a safe control policy. We now propose our combined loss function $\mathcal{L}_{combined}$ as shown in Equation~\ref{eq:loss_obj}, where the $w$ is a trade-off coefficient between discriminator loss and derivative loss. 
\begin{equation}
    \label{eq:loss_obj}
    \mathcal{L}_{combined} = \mathcal{L}_{policy} + w * \mathcal{L}_{derivative}
\end{equation}
By minimizing the $\mathcal{L}_{combined}$ via gradient descent, we could obtain a safer policy. The AIRL will be pre-trained to converge and provide a policy. The neural network $h_\omega$ pre-trained in Step 1 will be used to generate the set $\{s|h_\omega(s)\geq0\}$, where the states $s$ are explored by the policy.

\subsection{Data Collection}
\label{sec:data}
\vspace{-7pt}
To learn a CBF that enhances policy safety, we need to collect state sets $\mathcal{S}_s$, $\mathcal{S}_d$ before solving Equation~\ref{eq:optimization}. We assume the demonstration trajectories $\tau\in\mathcal{D}$ are safe and initialize $\mathcal{S}_s$ to be the set of states in the demonstrations. We cannot request demonstrators to take a risk of hurting themselves or damaging the robots to provide the dangerous states. Therefore, we define potentially dangerous states $\mathcal{S}_{pd}$ as a set that the agent has to pass before entering the $\mathcal{S}_{d}$. We design a new requirement ``\textbf{R2}$^\prime$: For $\forall s \in \mathcal{S}_{pd}$, $h(s) < 0$''. As stated in the preliminaries, the agent cannot enter set $\mathcal{S}_{pd}$ if \textbf{R1}, \textbf{R2}$^\prime$, and \textbf{R3} are satisfied. Hence, the agent cannot enter set $\mathcal{S}_{d}$ as well according to the definition of $\mathcal{S}_{pd}$. Therefore, \textbf{R2}$^\prime$ is a more strict requirement that prevents the agent from entering dangerous states, and we could take the \textbf{R2}$^\prime$ as \textbf{R2} and replace $\mathcal{S}_d$ in Equation~\ref{eq:y_func} with $\mathcal{S}_{pd}$. 

Defining potentially dangerous states ensures that CBF learns information about dangerous states while being safe for users. Therefore, demonstrators can safely provide potentially dangerous states to a set $\hat{\mathcal{S}}_{pd}$, which acts to be an approximation to the $\mathcal{S}_{pd}$. To avoid losing all the feasible paths to the goal, we will request demonstrators to try to collect states close to dangerous states. One example of a good potentially dangerous state close to dangerous state for a race car could be a position close to obstacles. In this work, we create $\hat{\mathcal{S}}_{pd}$ via code instead of real users. Specifically, 1024 states where the distance between the agent and the obstacle is less than a threshold are collected. We empirically show that the approximation $\hat{\mathcal{S}_{pd}}$ works well, as shown in Section~\ref{sec:result}.

\begin{table}[t]
\centering
% \footnotesize
\scriptsize
\begin{tabular}{ccclccc}
\hline
\multirow{2}{*}{Environment} & \multicolumn{2}{c}{Successful rate (Stdev)} &  & \multicolumn{3}{c}{Collision rate (Stdev)}      \\ 
\cline{2-3} \cline{5-7} 
& AIRL             & CBFIRL           &  & AIRL        & CBFIRL      & Improvement \\ \hline
2D racecar - 8 obstacles  & 0.97 (0.02)      & 0.95 (0.03)      &  & 0.58 (0.07) & 0.49 (0.09) & \textbf{15.52\%}     \\
2D racecar - 16 obstacles & 0.63 (0.28)      & 0.64 (0.07)      &  & 1.00 (0.12) & 0.80 (0.09) & \textbf{20.00\%}     \\
3D drone - 32 obstacles   & 0.34 (0.37)                 &    0.30 (0.33)              &  &      0.61 (0.40)       &      0.31 (0.19)       &       \textbf{49.18\%}      \\ \hline
\end{tabular}
\caption{This table shows the comparison between CBFIRL and AIRL on two domains.}
\label{tb:result}
\end{table}
\normalsize

\begin{figure}[t]
\centering
\begin{minipage}[t]{.45\textwidth}
  \centering
  \includegraphics[width=0.6\columnwidth]{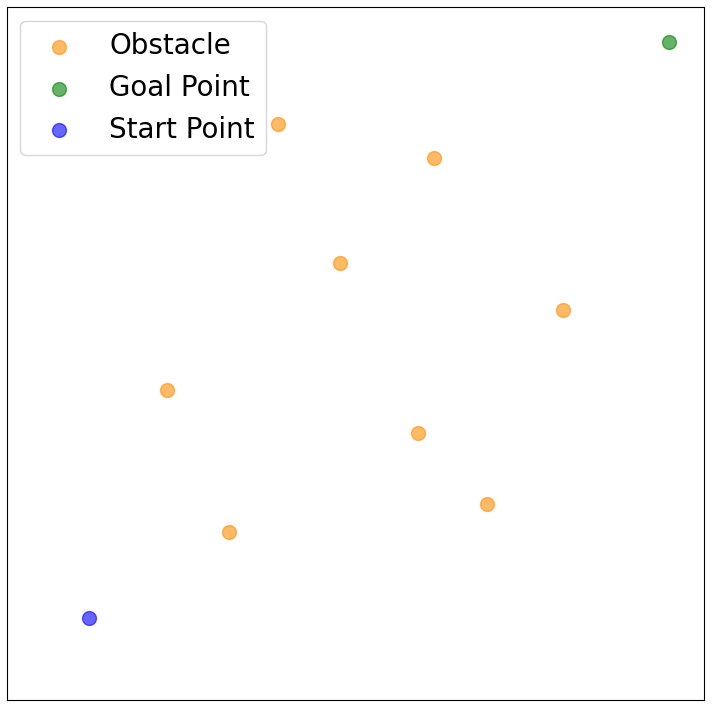}
  \caption{2D racecar environment.}
  \label{fig:env}
\end{minipage}
\hspace{.3cm}
\begin{minipage}[t]{.45\textwidth}
  \centering
  \includegraphics[width=0.8\columnwidth]{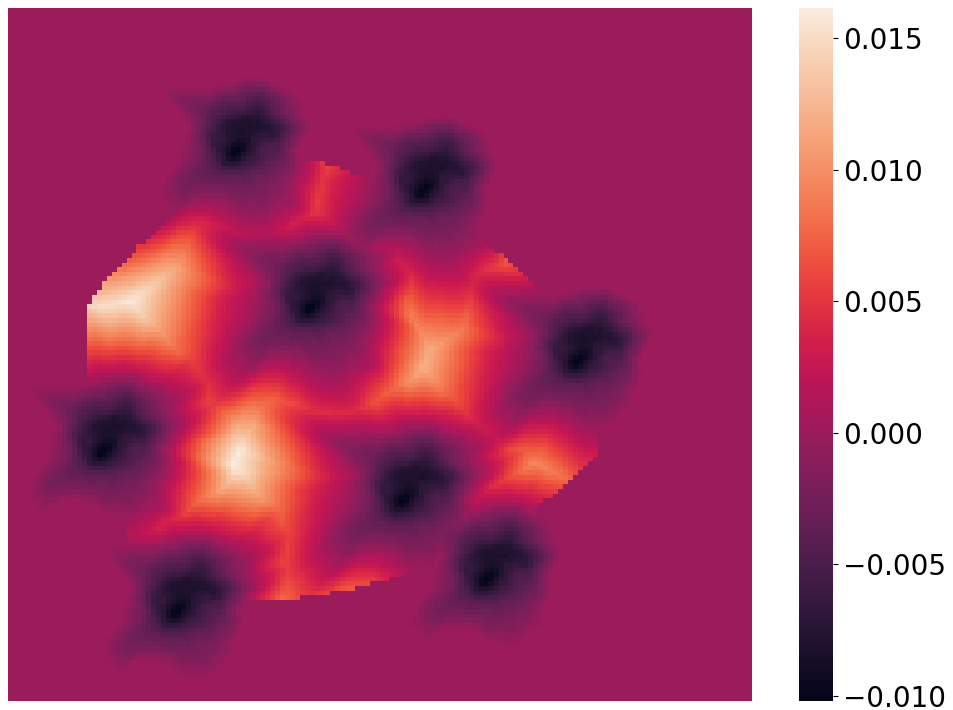}
  \caption{Heatmap of learned CBF.}
  \label{fig:heatmap}
\end{minipage}
\vspace{-10pt}
\end{figure}

\section{Experimental Results}
\label{sec:result}

% simulation envs - dynamics
% \subsection{Simulation Environments}
We evaluate CBFIRL on two simulated control environments: a 2D racecar and a 3D drone~\citep{qin2021learning}. For both environments, the agent travels across the map to reach the green target from the blue start point without colliding with the yellow obstacles that are moving. We define the state as the combination of the position and velocity of the agent and the nearest K obstacles. The episode terminates after 100 timesteps for the 2D racecar and 400 for the 3D drone. 
% We use the same dynamic model as~\citet{qin2021learning} and convert their multi-agent environment into single-agent setting.
We test two levels of difficulty in racecar (8 and 16 obstacles) and a setting of 32 obstacles for the drone domain. 

% For the racecar and drone, the model dynamics and the form of states for agent and obstacles are borrowed from

% As illustrated in Figure~\ref{}, we adopt two simulated control environments for evaluating the CBFIRL: 2D racecar and 3D drone. For each environment, the task is the same: the agile robot needs to reach the goal without colliding with obstacles.

% bar charts. 1. domain-1 for 8 & 16 obstacles 2. domain-2 for * & * obstacles
% \subsection{Results}
% \paragraph{Results --}
Two metrics are designed to evaluate the performance of the CBFIRL: ``Successful rate'' which measures the ratio of reaching the goal and ``Collision rate'' which shows the ratio of collision out of the 100 trajectories. We evaluate the two metrics on 100 trajectories to test CBFIRL's task success and safety against AIRL. We summarize the comparison in Table~\ref{tb:result}. Across the three environments, CBFIRL achieves a smaller collision rate than AIRL, which indicates a safer policy. Meanwhile, CBFIRL achieves a similar success rate as AIRL, which shows that our method has a good balance between safety and performance without being over-conservative to stand still.

% % \subsection{Combination between CBF and IRL}
% \label{sec:m1}
% \begin{figure}[t]
%     \centering
%     \includegraphics[width=0.7\textwidth]{imgs/vis.jpg}
%     \caption{The left Figure shows the 2D racecar environment with obstacles (yellow circles), the agent needs to cross the whole map from the start point (green circle) to the goal point (blue circle) without collision. The right Figure is the heatmap for control barrier function.}
%     \label{fig:vis}
% \end{figure}

% \begin{figure}%[t]
%     \centering
%     \subfloat[\centering label 1]{{\includegraphics[width=7cm]{imgs/real.jpg} }}%
%     % \qquad
%     \subfloat[\centering label 2]{{\includegraphics[width=9.3cm]{imgs/heatmap.png} }}%
%     \caption{2 Figures side by side}%
%     \label{fig:vis}%
% \end{figure}

% \begin{figure}
% \hfill
% \subfigure[Title A]{\includegraphics[width=5cm]{imgs/real.jpg}}
% \hfill
% \subfigure[Title B]{\includegraphics[width=5cm]{imgs/heatmap.png}}
% \hfill
% \caption{Title for both}
% \end{figure}

% visualization of h() and the training results (loss, acc for loss_barrier and loss_deriv)
% \subsection{Control Barrier Function Heatmap}
% \label{sec:cbf_heatmap}
% \paragraph{Control Barrier Function Heatmap --}
To evaluate the learned control barrier function $h(\cdot)$ in discriminating the safe set, $\mathcal{S}_s$, from the potentially dangerous states, $\hat{\mathcal{S}}_{pd}$, we visualize the $h(\cdot)$ for one 2D racecar state through the heatmap in Figure~\ref{fig:heatmap}. In generating the heatmap, We fix the positions of all obstacles and only move the agent over the map, which provides us with the corresponding $h(s)$ for varied $s$ to build the heatmap. Figure~\ref{fig:heatmap} show $h(s) < 0$ (darker) in the area where the agent is close to the obstacles (Shown in Figure~\ref{fig:env}) and provides qualitative evidence that the set $\hat{\mathcal{S}}_{pd}$ works well as an approximation of $\mathcal{S}_{pd}$.

%===============================================================================

\section{Conclusion and Future work}
\label{sec:conclusion}
\vspace{-7pt}
In this paper, we develop a novel framework, CBFIRL, to learn a safe policy from demonstrations by embedding the safety property of CBF into IRL methods. We transform the optimization problem of satisfying CBF conditions into a learning framework where the loss functions could be used to enhance the safety of IRL. We empirically validate that the proposed CBFIRL has fewer collisions for the two agile-robot domains than just IRL. In future work, we plan to extend CBFIRL to real robots. Another limitation of the current CBFIRL is that there remain some collisions for CBFIRL because of the balance between the safety and reachability objectives. Safety shield~\citep{alshiekh2018safe} can be a possible direction for future work.

% A possible direction for more intelligently collecting $\hat{\mathcal{S}}_{pd}$ can be active queries.

%===============================================================================

% \clearpage

% The acknowledgments are automatically included only in the final and preprint versions of the paper.
\acknowledgments{This work was sponsored by the National Institutes of Health (NIH) under grant number 1R56HL157457-01, Naval Research Laboratory (NRL) under grant number N00173-21-1-G009, and National Science Foundation (NSF) under grant number CNS-2219755.}

%===============================================================================

% no \bibliographystyle is required, since the corl style is automatically used.
\bibliography{ref}  % .bib

\end{document}